\documentclass{acm_proc_article-sp}

\begin{document}

\title{Plan or not: Remote Human-robot Teaming with Incomplete Task Information
}
%
%
%
%
%

%
\author{
}

\maketitle

\begin{abstract}

Human-robot interaction can be divided into two categories based on the physical distance between the human and robot: remote and proximal. 
In proximal interaction, the human and robot often engage in close coordination; 
in remote interaction, the human and robot are less coupled due to communication constraints.
As a result, providing automation for the robot in remote interaction becomes more important. 
Thus far, human factor studies on automation in remote human-robot interaction have been restricted to various forms of supervision,
in which the robot is essentially being used as a smart mobile manipulation platform with sensing capabilities. 
In this paper, we investigate the incorporation of general planning capability into the robot to facilitate peer-to-peer human-robot teaming,
in which the human and robot are viewed as teammates that are physically separated.
The human and robot share the same global goal and collaborate to achieve it. 
Note that humans may feel uncomfortable at such robot autonomy, which can potentially reduce teaming performance.
One important difference between peer-to-peer teaming and supervised teaming is that
an autonomous robot in peer-to-peer teaming can achieve the goal alone when the task information is completely specified.
However, incompleteness often exists, which implies information asymmetry.
While information asymmetry can be desirable sometimes, 
it may also lead to the robot choosing improper actions that negatively influence the teaming performance.
We aim to investigate the various trade-offs, e.g., mental workload and situation awareness, 
between these two types of remote human-robot teaming.

\end{abstract}



\keywords{Robot design principles, Autonomous robot capabilities, User study/Evaluation}

\section{Introduction}
\label{sec:introduction} 

There are two categories of human-robot interaction based on the physical distance between the human and robot: remote and proximal. 
While the human and robot often participate in close coordination in proximal interaction, 
they are less coupled in remote interaction due to communication constraints.
As a result, providing automation becomes more important for the robot in remote interaction. 

In remote human-robot interaction, since the human and robot are not in the same workspace, 
the human must remotely interact with the robot, and may have access to the sensor feeds (e.g., visual feeds from camera) on the robot.
So far, human factor studies on automation in remote human-robot interaction have been restricted to various forms of supervision,
in which the robot is essentially being used as a smart mobile manipulation platform with sensing capabilities. 
In supervised human-robot interaction, the human creates the plan to achieve the global goal,
and then either directly provides motion commands or breaks the plan into sub-plans (or sub-goals) for the robot to handle. 
During plan execution, the human can interact with the robot using a low level motion controller \cite{casper-tc-2008}, 
or a high level task and plan manager \cite{goodrich-smc-2003}.    
 
In this paper, we investigate the incorporation of general planning capability into the robot to facilitate peer-to-peer human-robot teaming,
in which the human and robot are viewed as teammates that are physically separated. 
In peer-to-peer teaming, the human and robot share the same global goal, and need to collaborate to achieve it.
Note that humans may feel uncomfortable at such robot autonomy, which can potentially reduce teaming performance.
One of the critical differences between our work and previous work (i.e., supervised teaming) is that, in our work, 
when the task information 
(e.g., environment settings, task specifications and other related knowledge) 
is completely specified {\em a priori} and modeled by the robot,
the robot can achieve the global goal alone, and potentially optimally.  
Note that this statement often does not hold in proximal human-robot teaming, 
in which the human may not only provide {\em knowledge} for teaming (which can be transferred in various forms through communication), 
but also physically participate in the execution (which may not be transferrable, e.g., special motor capabilities).
We aim to study the various trade-offs (e.g., mental workload and situation awareness) 
between peer-to-peer teaming and supervised teaming with remote human-robot interaction. 


However, task information often cannot be completely specified in real-world applications due to many reasons, 
which include dynamic environmental influences and practicality issues (e.g., one cannot build all related human knowledge into the robot).
For example, during an urban search and rescue task, 
the human teammate working outside of the disaster scene constantly receives updates from various sources, 
which need to be analyzed manually to extract useful information for the team;
the robot teammate working onsite also collects information for the team, while acting autonomously towards achieving the global goal. 
As a result, information asymmetry can be introduced between the human and robot, 
which can influence the teaming performance. 


In this paper, we perform investigation in remote human-robot scenarios, 
in which task information cannot be fully specified a priori, 
and the knowledge is not always synchronized between the human and robot.
This asymmetry may be due to the fact that certain task updates require manual processing, 
which is prone to interpretation delay 
(e.g.,  environment changes that can only be perceived and interpreted by humans),
or that the knowledge cannot be easily incorporated into the planning model (e.g., inferencing and image processing capabilities of humans).
While information asymmetry can be detrimental sometimes, it can also be desirable from the point of view of human cognitive load
(e.g., the human does not need to know how many rooms the robot travels through to move to a specified location). 

We choose the urban search and rescue (USAR) task \cite{nourbakhsh-pc-2005} to conduct our investigation.
USAR tasks represent remote human-robot interaction scenarios in which task information is often incomplete 
(e.g., due to environmental changes caused by disasters).
In an USAR task, the human-robot teams are deployed during early phases of an emergency response. 
The aim is to explore areas of the disaster scene and provide real-time information,
which is then used for situation assessment to aid emergency management of the rescue team 
(e.g., places to be avoided).
While the human and robot teammates constantly interact to collaborate,
the human may also need to coordinate with other operators and related personnel. 
Both objective (e.g., task performance) and subjective measures (e.g., immediacy) 
are to be collected in this task to establish the trade-offs between peer-to-peer and supervised teaming. 

\section{Related Work}
\label{sec:related-work}

Many research works have been conducted on proximal human-robot interaction. 
Given the proximity between the human and robot, proximal interaction tends to be rich and multimodal. 
For example, previous research has investigated how speech (e.g., \cite{scheutz-hri-2006}), 
gaze (e.g., \cite{staudte-hri-2009}) and gesture (e.g., \cite{nickel-image-2007}) are used,
as well as their relationships during the interaction (e.g., \cite{stiefelhagen-iros-2004}). 
Furthermore, researchers have also studied how humans can be considered during proximal interaction 
for safety reasons (e.g., \cite{sisbot-tro-2007}), 
as well as how human preference influences interaction (e.g., \cite{unhelkar-hri-2014}).

Meanwhile, in remote human-robot interaction, the human and robot are often less coupled.
Given the communication constraints (e.g., communication bandwidth, reliability and modalities) due to the physical separation,
providing automation for the robot in remote interaction becomes more important.
However, so far, human factor studies on automation in human-robot teaming have been mainly restricted to various forms of supervision,
in which the human decides the actions \cite{ruff-ptve-2002} or sub-goals \cite{heath-hpsa-2004} for the robot towards achieving the global goal.
While the robot is allowed to propose plans for the sub-goals,
the human must interact with the robot after it achieves the current sub-goal.
During the missions, the human interacts with the robot either through motion controllers \cite{casper-tc-2008,goodrich-issr-2007}, 
or task and plan managers \cite{goodrich-smc-2003}.
A comprehensive review of related works on supervisory human-robot teaming can be found at \cite{chen-tsmc-2011}.

In this paper, we discuss peer-to-peer human-robot teaming in which the human and robot are viewed as teammates.
The human and robot share the same global goal. 
As a result, when the task information is completely specified, the robot can achieve the task alone.
With incomplete task information, the human and robot must handle potential information asymmetry.  
While the human teammate may receive updates \cite{cantrell-hri-2012} about the environment (e.g., from other operators), 
or have knowledge that the robot teammate does not have (e.g., image processing and analyzing capabilities), 
the robot can observe and interact directly with the real environment (e.g., detecting that a door is blocked).
The human and robot teammates may interact to resolve information asymmetry during the task. 
While there are works that incorporate general planning capability into robots to achieve peer-to-peer human-robot teaming (e.g., \cite{talamadupula-tist-2010}), 
there exists no empirical investigation on its influence on human-robot teaming performance and the trade-offs. 

Regarding the benefits of automation in human-robot interaction, 
it is well known that automation can have both positive and negative effects on human performance.
Empirical proofs have been provided in four main areas: mental workload, situation awareness, complacency and skill degradation \cite{parasuraman-erg-2000}.
Many earlier human factor studies have been on how automation should be designed to benefit human-machine interaction \cite{visser-cedm-2011}. 
Researchers have also characterized the types \cite{parasuraman-erg-2000} and levels of autonomy \cite{parasuraman-smc-2000}.
Regarding efficient human-machine teaming, it is argued in \cite{christoffersen-hpce-2002} that automation should be provided in the context of interaction. 
Along this line, a substantial amount of work has investigated how efficient teaming can be achieved \cite{klein-tcm-2004,sheridan-rhfe-2005}, 
which is also applicable to human-robot teaming scenarios. 
We build our human-robot interaction interface based on these previous works.

\subsection{Background}
\label{subsec:background}

When task information is completely specified {\em a priori}, a task can be compiled into a problem instance for an automated planner. 
A plan is created by connecting initial state to the goal state using agent actions. 
A planning problem can be specified using a planning domain definition language (PDDL) \cite{mcdermott-pddl-1998}.
Depending on the task, there are many extensions of PDDL (e.g., \cite{fox-jair-2003,gerevini-pddl3-2006}) that incorporate various modeling requirements.
In this paper, we use the extension of PDDL described in \cite{fox-jair-2003} to model an USAR task.
While using an automated planner allows an automated agent to reason directly about the global goal, 
the limitation comes from the completeness assumption.  

Regarding this assumption, 
while completeness often cannot be guaranteed (e.g., dynamic environmental changes cannot be modeled or predicted), 
there are various types of incompleteness that can be considered in an automated planner.
For example, partial satisfaction of the goal \cite{briel-psp-2004}, incomplete initial state \cite{Hoffmann2006507}, 
incomplete action models (i.e., used to perform planning) \cite{kambhampati-lite-2007}, 
as well as incomplete preference of the human teammate \cite{nguyen-partialp-2012}.

\subsection{Hypotheses}
\label{subsec:hypotheses}

In this paper, we aim to investigate the incorporation of an automated planner to achieve remote peer-to-peer human-robot teaming.
We compare peer-to-peer teaming, in which the robots have a general planning capability, with supervised teaming, in which the capability is absent. 
In particular, we make the following hypotheses:

$H1)$ Robot with a planning capability provides more natural teaming experience.

$H2)$ Robot with a planning capability reduces human teammate's mental workload; on the other hand, it also reduces situation awareness.

$H3)$ Robot with a planning capability gradually reduces the interaction between the human and robot. 

\section{Study Design}
\label{sec:design}

Our study is focused on comparing robot teammate with (i.e., peer-to-peer) and without (i.e., supervised) a planning capability in human-robot teams that assist in making situation assessments during urban search and rescue (USAR) tasks. 

\subsection{Environment}

\begin{figure}
\centering
\epsfig{file=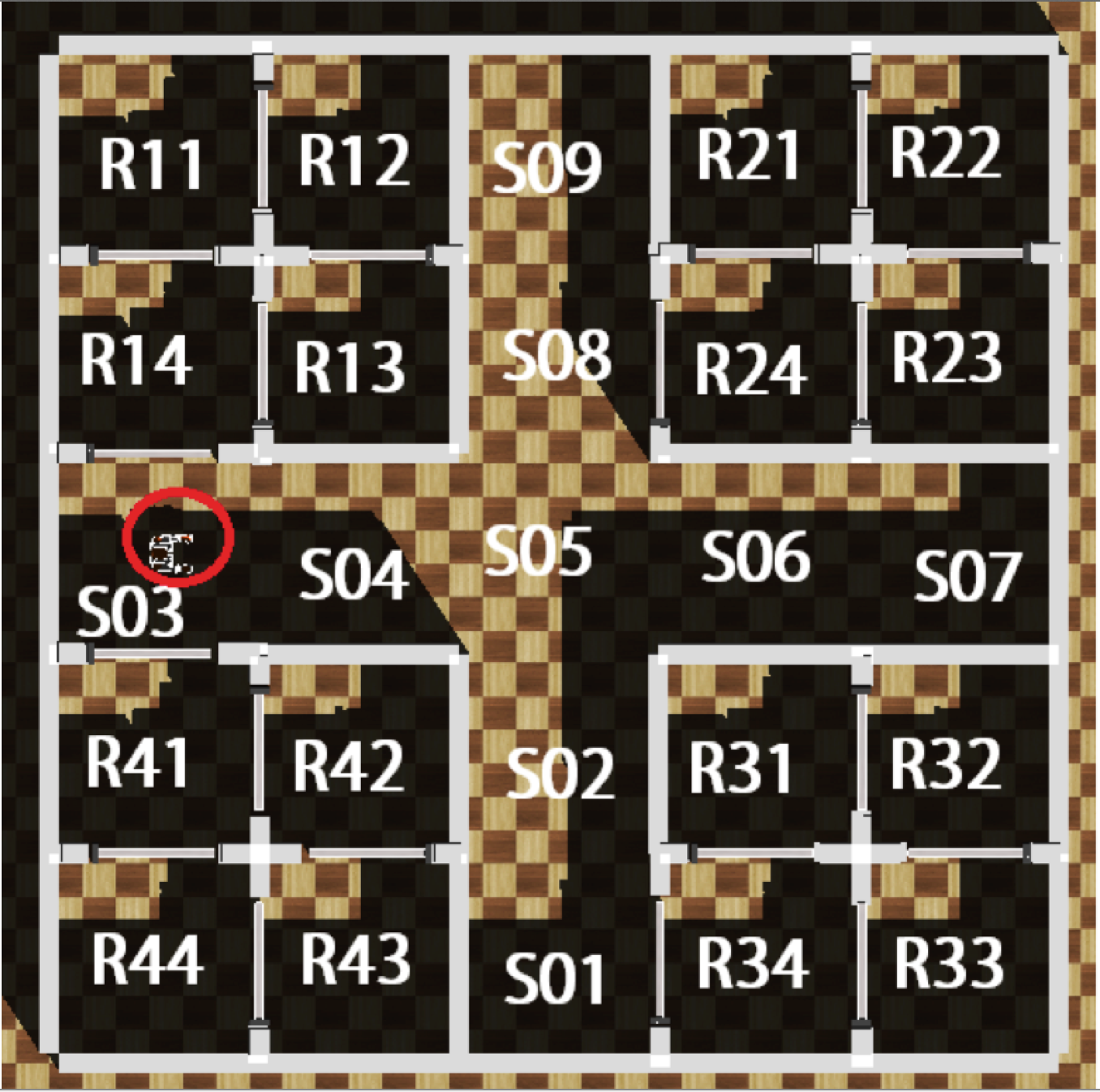, height=2in, width=2in}
\caption{Environment used in the USAR task with a simulated Nao humanoid robot}
\label{fig:environment}
\end{figure}

Fig. \ref{fig:environment} is the simulated environment (created in Webots) used in our USAR task,
which represents the floor plan of an office building before a disaster occurs (e.g., a fire). 
The environment is organized as segments, and each segment is identified by a unique label (e.g., $R11$) in Fig \ref{fig:environment}.  
Furthermore, the segments that represent rooms are grouped into four regions, with each region containing four rooms.
For example, one of the regions, denoted by $R1$, contains four rooms $R11, R12, R13$ and $R14$. 
Each region can be accessed via a door that connects to a hallway segment and the rooms in each region are also connected by doors. 
The doors are initially closed and can be pushed open by the robot. 
The doors remain open after being pushed open.
The robot teammate works inside this environment and the human teammate remotely interacts with it.  
The environment is designed such that no significant computational skills are required, 
in order to remove the influence of these skills on the performance.


\subsection{Task Specification}

\begin{figure}
\centering
\epsfig{file=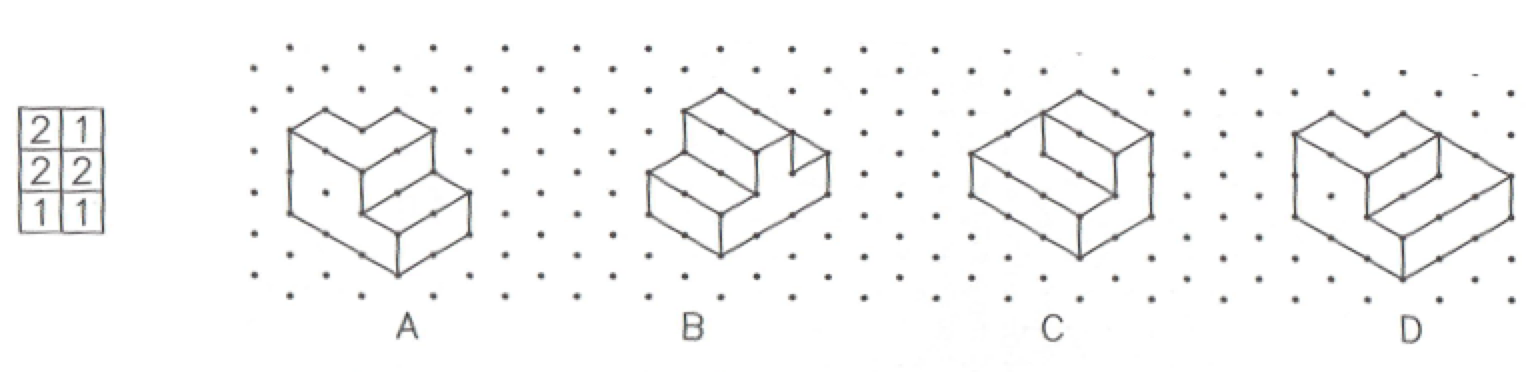, height=0.8in, width=3.3in}
\caption{Example question of the secondary task}
\label{fig:secondary-task}
\end{figure}

The global goal of this USAR task is to report the number of casualties in as many rooms as possible, given a certain amount of time.
This is to simulate that the human-robot teams only have a limited amount of time to perform the situation assessment in a disaster response scenario.
The robot starts in the position that is specified by a red circle in Fig. \ref{fig:environment}.
In this task, both the human and robot have access to the floor plan before the disaster,
and hence both can independently determine the ordering in which they plan the rooms to be reported. 

Meanwhile, the incomplete task information can be a result of blocked doors and other environmental changes. 
To simulate this incompleteness, we design the doors inside the regions (which connect adjacent rooms) in such a way that some of them are blocked.
Note that the room with a blocked door would be accessible by the other door, so that it would still be of interest to report it.  
We provide the information (regarding which doors may be blocked) to the human teammate {\em after} the task starts to simulate information coming from other sources (e.g., monitoring cameras or other human operators).
The same information remains unknown to the robot teammate, which represents information that is inaccessible to the robot teammate.  
Although the robot can learn this information by pushing the door and failing, this can reduce the teaming performance. 

While the human teammate needs to interact with the robot to mitigate the influence of the information asymmetry, 
the human must also process and analyze information from the robot teammate and other sources (e.g., other human operators), 
such that the human may not have time to micro-manage the robot.
To simulate this, the human is also assigned to a secondary task.
This secondary task involves solving three-dimensional visualization puzzles (see Fig. \ref{fig:secondary-task} for an example).
These puzzles require only basic visualization skills and reasoning abilities.
The performance of the team is evaluated on both the primary and secondary tasks.

\subsection{Interface Design}

\begin{figure}
\centering
\epsfig{file=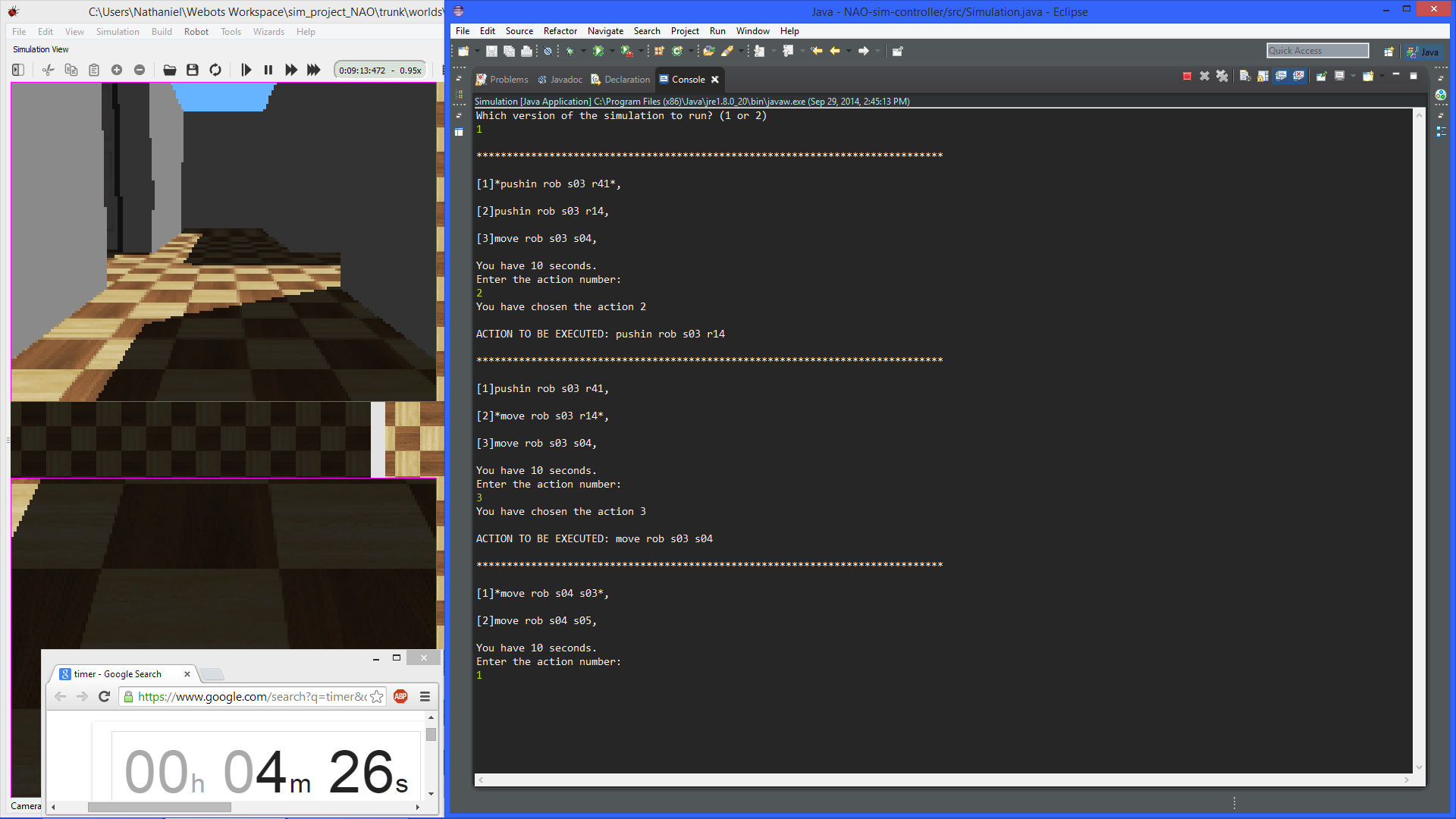, height=1.65in, width=3.3in}
\caption{Interaction Interface in the USAR task}
\label{fig:interface}
\end{figure}

To create a more realistic USAR environment, the human teammate only has access to the visual feeds from the robot teammate.
In other words, the human can only observe the part of the environment from the robot teammate's eyes (i.e., cameras).
The interaction interface with the robot is shown in Fig. \ref{fig:interface}.

The interaction interface between the human and robot teammates are the same for the robot with or without a planning capability. 
In both cases, the robot teammate would display a list of applicable actions that it can perform given the current state.
However, note that the level of automation \cite{parasuraman-smc-2000} that can be supported varies between these two cases. 
In the first case (robot with a planning capability), the robot teammate knows exactly the sequence of actions to achieve the global goal;
in the second case, the robot teammate can only filter out actions that are not applicable given the current state.
To limit the performance difference due to the level of automation, 
we set the level of automation to be the lower bound that is applicable to the first case, 
while setting it to be the upper bound for the second case,
according to \cite{parasuraman-smc-2000}. 
In our study, the two cases correspond coarsely to level $6$ (management-by-exception) and $3$ described in \cite{parasuraman-smc-2000}, respectively. 
Note that level $10$ represents full autonomy and level $1$ is manual operation. 

In particular, in the first case, while the robot provides the list of applicable actions given the current state, 
it also highlights the next action in the plan (for achieving the global goal), which is selected by the robot using its planning capability.
The robot computes this action by creating a plan that satisfies all {\em remaining} goals (each room that is not reported is a remaining goal). 
Note that the next action in the plan must be in the list of applicable actions. 
Although the robot's next action is highlighted, 
the human teammate is free to interact with the robot teammate to choose any action from the list of applicable actions. 

Every time that an action is completed by the robot, the interaction interface alerts the human teammate (i.e., by playing a beeping sound), 
and provides $10$ seconds for the human to interact with the robot.
If the human does not interact with the robot within $10$ seconds, 
in the first case, 
the robot executes the next action in its plan;
in the second case, the robot simply continues to wait.  
If the human teammate interacts with the robot, 
the robot teammate would execute the action chosen by the human teammate. 
This process is repeated until the given time elapses. 

\subsection{Experimental Setup Process}

\begin{figure}
\centering
\epsfig{file=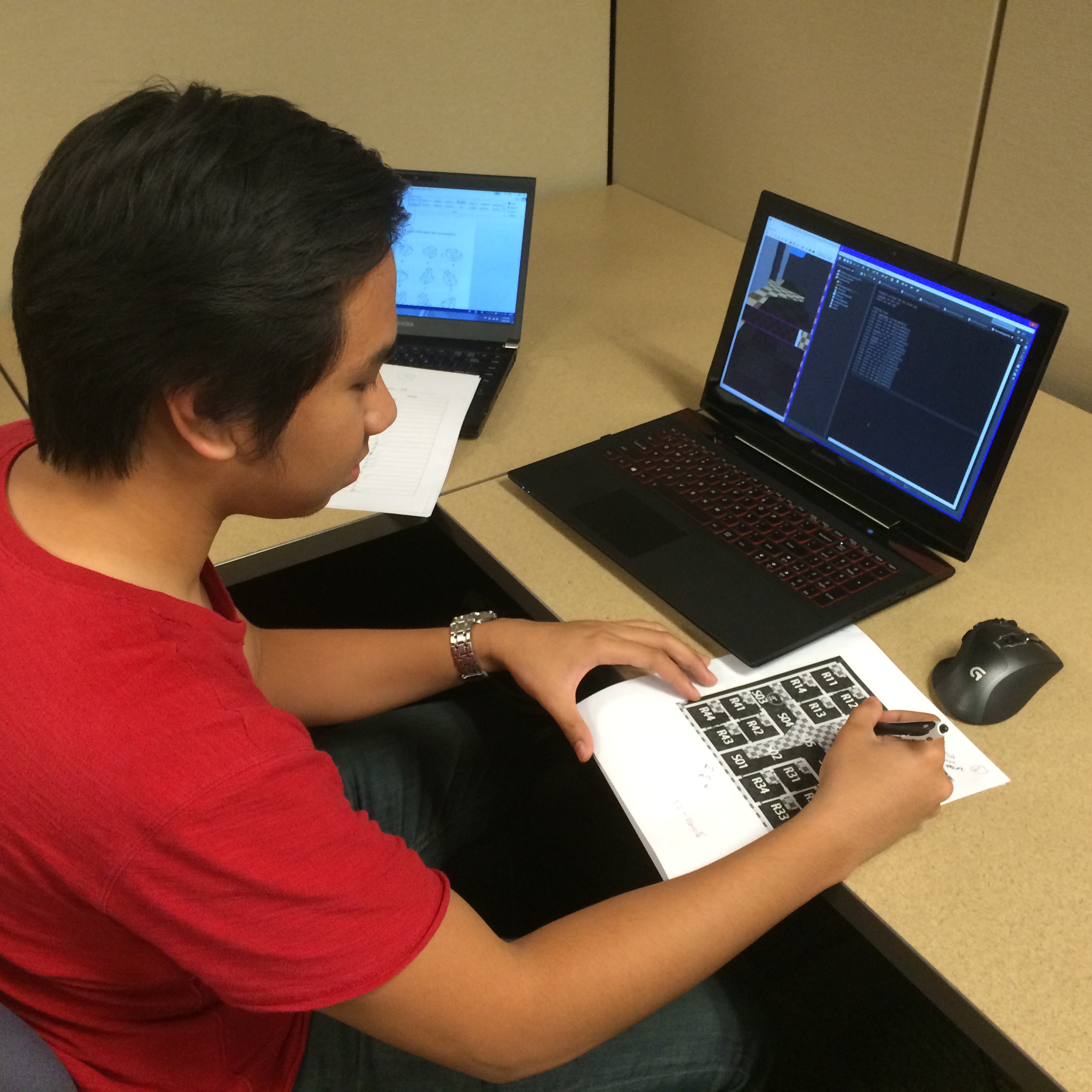, height=2in, width=2in}
\caption{Experimental setup in the USAR task}
\label{fig:setup}
\end{figure}

The experiment was set up in our lab space.
Participants were assigned alternatively to team up with either a robot with a planning capability or a robot without in the USAR task. 
Each participant is only allowed to take part in one experimental trial to avoid performance fluctuation due to experience. 
All participants completed the consent form before participating in the study. 
Prior to each run, the participant was asked to read the instruction materials.
The participant was then exposed to the simulator and the interface, 
and was asked to experiment with them to gain some familiarity. 

During each of the actual trials, a map of the simulation environment (as shown in Fig. \ref{fig:environment}) was provided to the participant,
in order to provide the situation awareness of the initial state (i.e., robot initial position in the map). 
The participant was also provided with the secondary task questions and a separate answer sheet to fill in the responses.  
The participant was asked to collaborate with the robot to report as many rooms as possible within 20 minutes. 
In addition to interacting with the robot teammate to report rooms, 
the participant was also instructed to work on the secondary task, 
and was informed that the teaming performance would be evaluated based on both the primary and secondary tasks.
After one minute into the run, the extra information (regarding which doors were blocked) was given to the participant.
Each trial ended when the given time elapsed.
Finally, the participant completed the questionnaire (in Likert scale) that was designed to qualitatively evaluate various human-robot interaction aspects.  
Fig. \ref{fig:setup} shows the experiment setup in one of the trials. 

\section{Results}
\label{sec:results}

The study was performed over $4$ weeks and involved $19$ volunteers ($11$ males, $8$ females),
Volunteers have ages with $M = 24.47$ and $SD = 1.07$.
These participants were recruited from students on campus. 
Due to the requirement of understanding English instructions, 
participants must indicate that they are confident with English communication skills before taking part in the study. 
We also asked about the participants' familiarity with computers ($M = 6.68$, $SD = 0.48$), robots  ($M = 2.74$, $SD = 0.73$),
puzzles for the secondary task ($M = 3.58$, $SD = 1.02$), 
and computer gaming ($M = 3.79$, $SD = 1.18$), in seven-point scales 
(with $1$ being least familiar and $7$ being most familiar) after the study. 
The participants reported familiarity with computers, 
but not so much with robots, puzzles for the secondary task or computer gaming.

\subsection{Measurement}

A post-study questionnaire is used to evaluate three of four areas that are often used to assess automated systems:
mental workload, situation awareness, and complacency \cite{parasuraman-erg-2000}.
Furthermore, we also use the questionnaire to evaluate several psychological distances between individuals,
which include immediacy, effectiveness, likability, and trust of the robots.
Immediacy describes the participant's feeling about how engaging the robot is.
Effectiveness describes the participant's feeling about how effective the robot is as a teammate.
Likability describes how likable the participant feels about the robot.
Trust describes whether the participant feels that the robot is trustworthy.
We also collect participants' opinions on whether they think that the robot should be improved (i.e., {\em improvability}).

One-way fixed-effects ANOVA tests were performed to analyze the objective performance and measures,
as well as the subjective questions.
The fixed factor in the tests is the type of the robot, 
which is either a robot with a planning capability (i.e., peer-to-peer or P2P) or without (i.e., supervised).


\subsection{Objective Performance}

We first investigate the objective performance and measures. 
The performance of the human-robot teaming is evaluated both on the primary task and secondary task.
The primary task is evaluated based on the number of rooms reported in the $20$ minutes given.
The second task is evaluated based on the number of puzzles that the participant answered correctly and incorrectly.

\begin{figure}
\centering
\epsfig{file=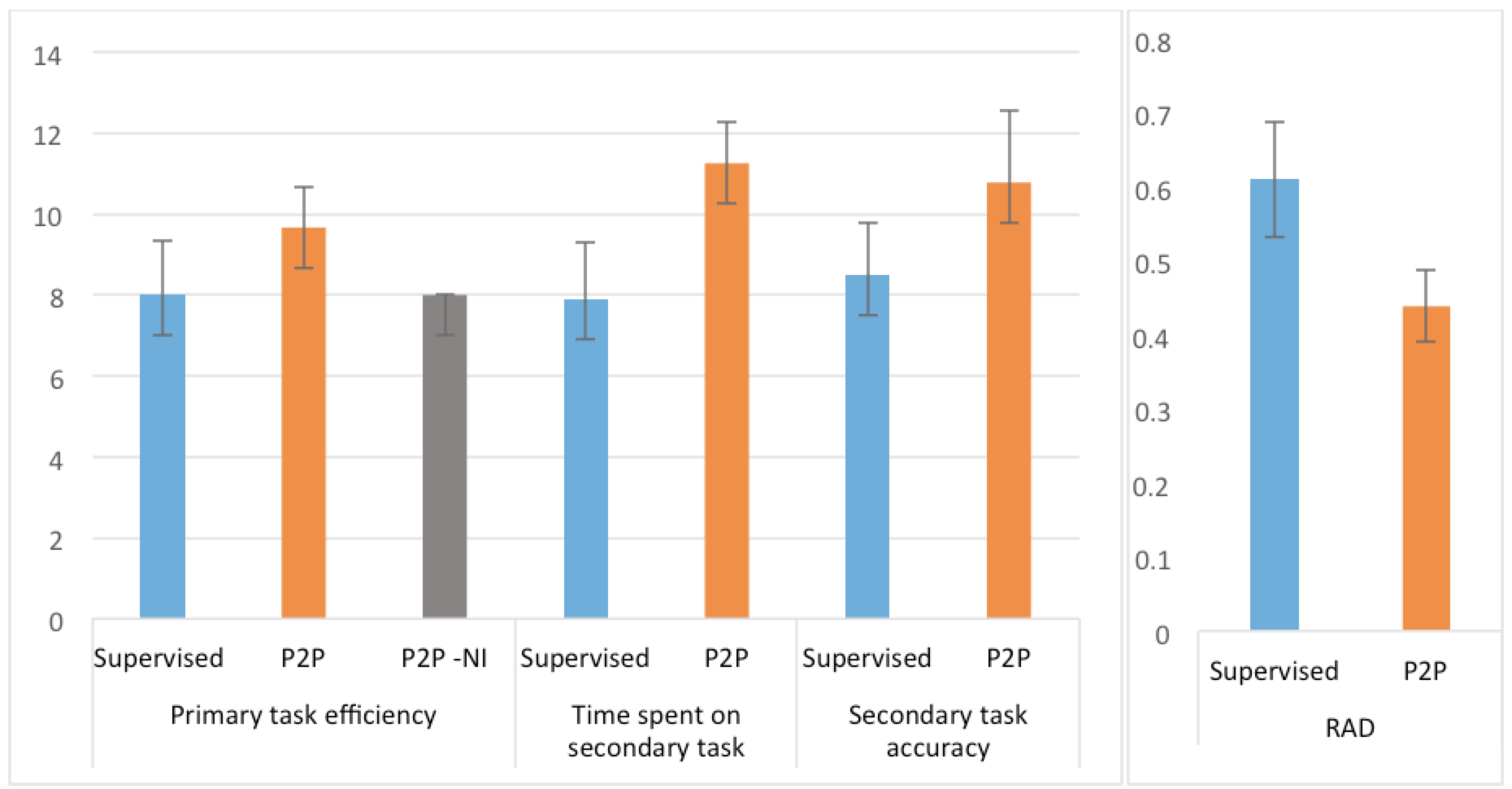, height=1.6in, width=3.3in}
\caption{Results for objective performance and measures.}
\label{fig:obj}
\end{figure}

\subsubsection{Primary Task}

For evaluating the primary task performance, we add a third type of robot: 
robot with a planning capability but without interaction (P2P-NI). 
Note that a P2P-NI robot can still achieve the task alone in our USAR scenario,
although the performance would be degraded due to the incomplete task information.
We ran the P2P-NI robot $3$ times for $20$ minutes each.
The variance for the runs is small.
Comparison with this type of robot is used (as a baseline) to determine whether human-robot teaming with a planning robot improves efficiency over planning robot performing alone. 
The analysis of variance shows a significant difference on the performance metric for the primary task, $F(2, 19) = 19.56, p < 0.001$.
The result is also presented in Fig. \ref{fig:obj}.
We can also see that the robot without a planning capability did not perform better than the P2P-NI robot performing alone,
but the variance with interaction is larger in our USAR task. 

\subsubsection{Secondary Task \& Mental Workload}

We evaluated the secondary task performance based on both the time spent on the task and the accuracy.
To discourage participants from guessing the answers to the puzzle questions, 
they were told that each incorrect answer would cost them half of the score of the question 
(i.e., taken out of the scores they obtained from correct answers).
Furthermore, we also evaluated the robot attention demand (RAD) \cite{Crandall-iros},
which is the percentage of the participant's time dedicated to human-robot interaction. 
RAD is often used as an indicator for mental workload. 
The results are also presented in Fig. \ref{fig:obj}.
Our analysis shows significant differences on these performance metrics and measures,
$F(1, 17) = 10.43, p < 0.01$ for secondary task accuracy, $F(1, 17) = 32.35, p < 0.001$ for time spent on the secondary task, 
and $F(1, 17) = 9.35, p < 0.01$ for RAD.

Another note is about the relationship between RAD and Fan out \cite{Crandall-iros}. 
Fan out determines the maximum number of robots that a human can simultaneously interact wth. 
In our USAR task scenario, we have this measure for robot with a planning capability and without as $2.30$ and $1.66$, respectively,
showing that it is easier to manage robots with a planning capability. 

\subsubsection{Skipped Actions}

\begin{figure}
\centering
\epsfig{file=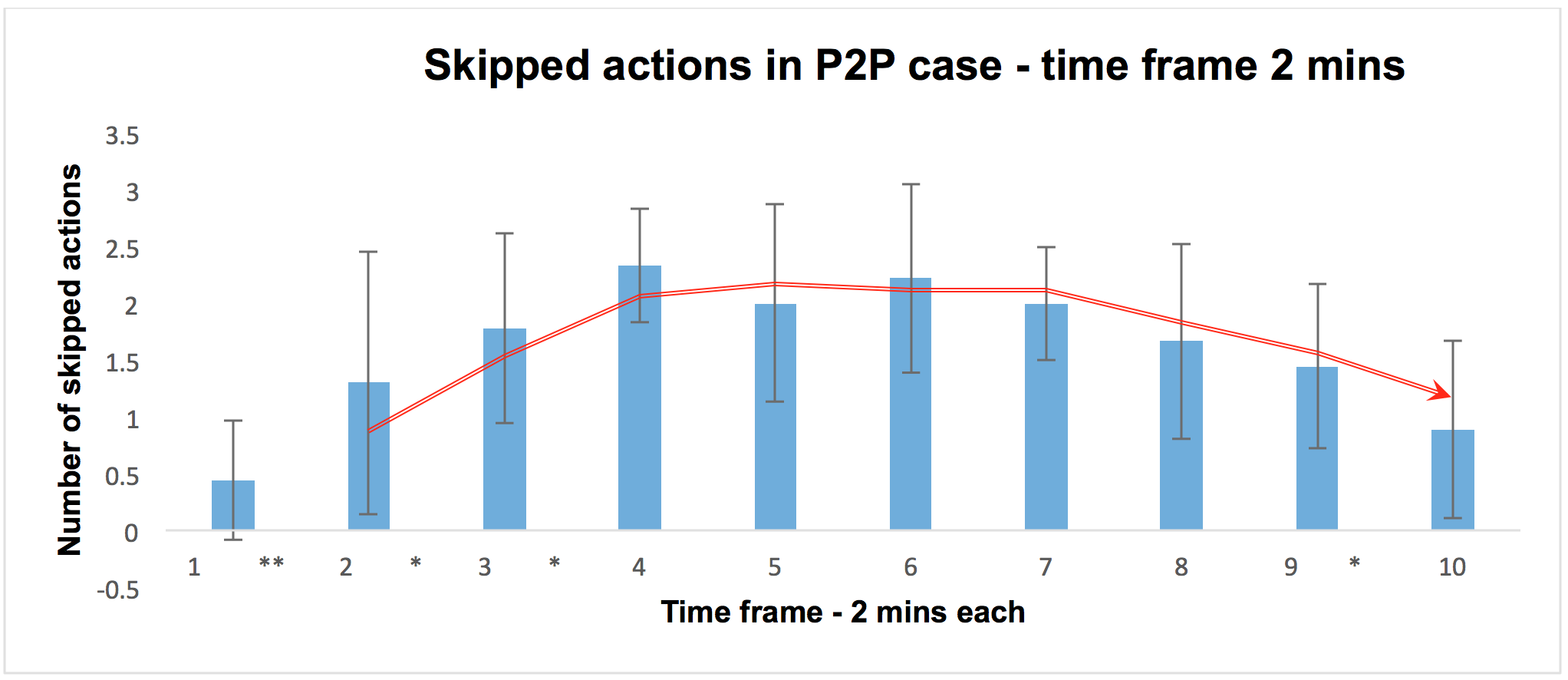, height=1.9in, width=3.4in}
\caption{Results for skipped actions. We split the task execution into 2-minute time frames, and tested the differences between any two consecutive frames.
$*$ denotes $p < 0.05$, $**$ denotes $p <  0.01$.}
\label{fig:skipped}
\end{figure}

The {\em skipped actions} measure is computed as the number of actions that the human teammate failed to interact with the robot with a planning capability within $10$s.
In such cases, the robot would execute the next plan action.
Here, we test our hypothesis ($H3$) about the change of interaction time during the task. 
For supervised human-robot teaming, given that the human needs to constantly guide the robots via action or sub-goal selection, 
the RAD would remain relatively stable. 
However, when the human teammate works with a robot that understands how to (accurately) perform the task by itself 
(testing of inaccurate robot teammate is future work), 
we anticipate that the human teammate would learn about this and gradually reduce the interaction time (thus increasing the skipped actions). 
This may also be one of the most significant differences between peer-to-peer teaming and supervised teaming.

On the other hand, similar to the fact that high level automation can reduce situation awareness (or even cause skill degradation), 
we also anticipate the decrease of situation awareness ($H2$). 
This hypothesis is investigated in subjective measures following next.
Fig. \ref{fig:skipped} shows the change of skipped actions during the task execution, split into $2$-minute time frames. 
Our analysis shows significant differences between the consecutive time frames using student's t-test. 
Fig. \ref{fig:skipped} is partially consistent with $H3$ until approaching the end the task,
in which the number of skipped actions reduces (thus interaction increases). 
This shows that the human teammate preferred to interact more as the time ran out, 
potentially as an approach to improving performance.

\subsection{Subjective Performance}

\begin{figure*}
\centering
\epsfig{file=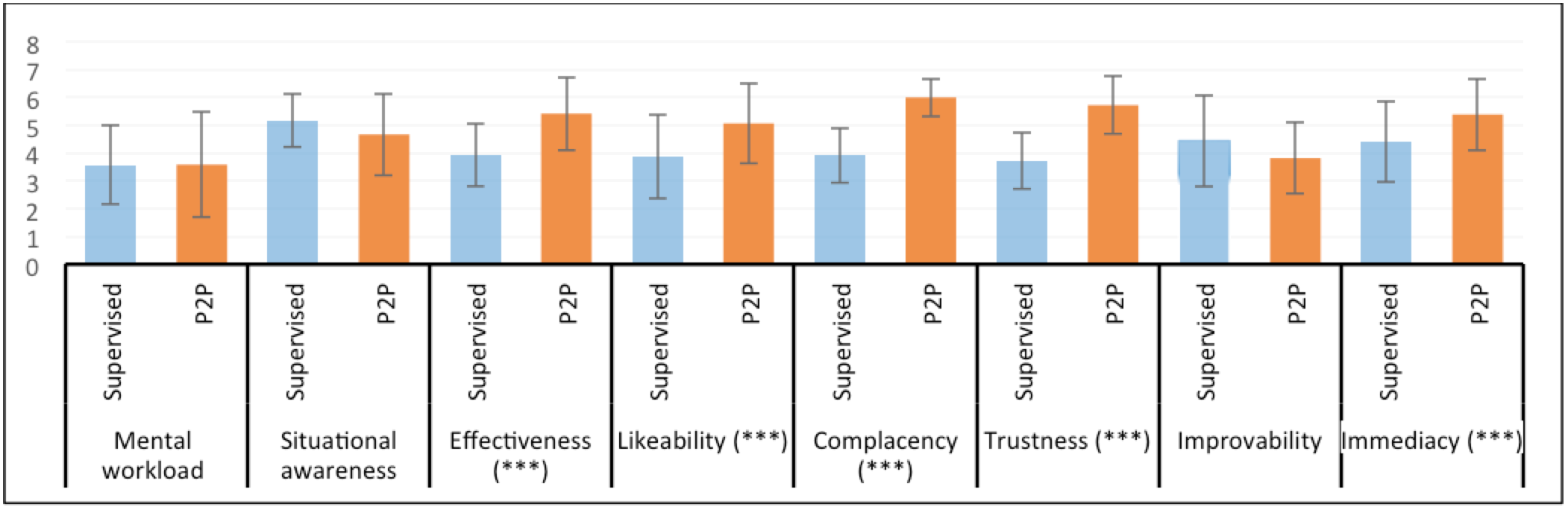, height=2.0in, width=6.2in}
\caption{Results for subjective measures. $*$ denotes $p < 0.05$, $**$ denotes $p <  0.01$, $***$ denotes $p < 0.001$.}
\label{fig:sub}
\end{figure*}

In this section, we investigate the subjective performance based on the questionnaire ($17$ questions in total).
For these $17$ questions, we categorize them into $8$ different (partially overlapping) groups: 
mental workload ($4$ items, Cronbach's $\alpha = 0.71$), 
situation awareness ($3$ items, Cronbach's $\alpha = 0.69$), 
complacency ($2$ items, Cronbach's $\alpha = 0.75$).
Furthermore, we also use the questionnaire to evaluate several psychological distances between the human and robot teammates,
which include 
immediacy ($4$ items, Cronbach's $\alpha = 0.86$), 
effectiveness ($4$ items, Cronbach's $\alpha = 0.77$), 
likability ($3$ items, Cronbach's $\alpha = 0.70$), 
and trust ($4$ items, Cronbach's $\alpha = 0.82$) of the robots.
We also include the improvability ($3$ items, Cronbach's $\alpha = 0.69$).
The answers to the questions are in seven-point scales.  
The results are presented accumulatively in Fig. \ref{fig:sub}.

\subsubsection{Mental Workload}

For mental workload, we include questions that inquire about the ease of working with the robot, 
and questions to rate the participant's workload to interact with the robot.  
Our analysis does not find any significant difference ($p = 0.98$). 
This is an interesting result:
although the participants working with a robot with a planning capability experienced increased
teaming performance, the participants did not feel any difference about the mental workload, 
when compared to participants working with a robot without a planning capability. 
This result is inconsistent with $H1$, which may be due to following reason. 
Given that the participants were not told to follow the plan actions selected by the robot with a planning capability, 
it seemed that most humans would prefer to rely on themselves at the beginning nonetheless. 
This may be a result of the lack of trust in the robot for handling the task initially. 
This effect also seems to be consistent with that shown in Fig. \ref{fig:skipped}.

\subsubsection{Situation Awareness}

For situation awareness, we include questions that inquire about the awareness of the positions of the robot,
and the understanding of the robot actions during the task.
Our analysis does not show a significant difference ($F(1, 17) = 3.50, p = 0.67$). 
This result is inconsistent with $H2$.
This may be partially due to the fact that the selected action of the robot during the task execution provided situation awareness to the human teammate,
since the same action is also likely to be chosen by the human in the same situation, given the common goal. 

\subsubsection{Complacency}

For complacency, we include questions about the comfort and ease of the teaming. 
Our analysis shows that $F(1, 17) = 56.82, p < 0.001$.
This is consistent with the objective performance and measures, and $H1$, 
which shows that the human generally feels more satisfied working with a robot with a planning capability. 

\subsubsection{Immediacy, Effectiveness, Likability \& Trust}

For immediacy, we include questions about how useful the participant felt about the interaction,
about the robot as a teammate, 
and how much the participant felt that the robot shared the same motivation and common goal. 
Our analysis shows that $F(1, 17) = 13.17, p < 0.001$.

For effectiveness, we include questions about the perceived effectiveness of the team, the robot,
and whether or not the participant felt that the robot sometimes performed unexpectedly.
Our analysis shows that $F(1, 17) = 33.64, p < 0.001$.

For likability, we include questions about both the ease and comfort, 
whether the participant felt that the interaction between them was effective, 
and whether the robot took initiatives to achieve the common goal.  
Our analysis shows that $F(1, 17) = 28.92, p < 0.001$.

For trust, we include questions about the evaluation of the robot performance,
and how much the participant felt that they worked as a real team (instead of a supervised team).  
Our analysis shows that $F(1, 17) = 71.57, p < 0.001$.
These results are consistent with $H1$.

\subsubsection{Improvability}

For improvability, we include questions about how much the participant felt that the robot could be improved, 
and how the participant evaluated the robot performance and interaction. 
Our analysis does not show a significant difference for improvability ($p = 0.06$).
The reason for this result could be that there were no expectations from the participants on how the robots
should perform in such tasks, given their unfamiliarity with the task and robots.

\subsection{Summary}

In summary, our results are partially consistent with our hypotheses. 
In general, the participants seemed to prefer to engage in peer-to-peer teaming and the task performance was higher than supervised teaming.
However, our expectation is that peer-to-peer teaming may not reduce mental workload in short-term tasks. 
This is understandable since working with a proactive teammate may require more interactions at the beginning.
Once a mutual understanding is established, the mental workload may be reduced.
In our USAR task, given the short span of our task, the reduction is not prominent. 

The result on situation awareness is also interesting.
Given that the situation awareness is not reduced significantly in peer-to-peer teaming, 
it seems to suggest that providing ``recommended actions'' may also be used as a way to reduce the negative influence of automation. 
For example, if the human teammate is being exposed to both high and low level robot actions during the task (rather than sub-goals as in supervised teaming),
the situation awareness may potentially be maintained. 
This is intuitive in human-human interaction, since human teammates may only choose to inform others about the details of actions that they are uncertain of. 
This is also more broadly related to research directions in plan explanation and excuse generation.

\section{Limitations and Future Work}
\label{sec:limitations}

One of the most obvious limitations of this work is the simplicity of the planning domain model,
which is a result of the simplification of the USAR task scenario that we considered. 
In particular, the USAR task scenario that we use to test our hypotheses involves only three types of actions:
$Move$, $Push$ and $Report$. 
Although these actions can be considered as abstract actions that involve complex action sequences, 
real-world USAR and other scenarios often involve significantly more actions.
This can cause delays for the planning capability on the robot, which need to be investigated.
On the other hand, due to the increased complexity, 
we may also expect the planning system to offer more advantages than supervised teaming.
This is especially true when sub-goals cannot be simply identified 
(e.g., when one needs to decide the best visiting sequence of locations to ensure timely coverage in a complex environment). 
Meanwhile, proper plan explanation and excuse generation methods may need to be incorporated,
 given the difficulty of complex plan understanding.
 Furthermore, this difficulty may also come from the fact that the task information is incomplete (hence there is information asymmetry). 
We plan to investigate these factors for peer-to-peer human-robot teaming in future work.
 
Another obvious limitation is the influence of automation level in our analysis:
in particular, the level of automation for supervised human-robot teaming in our analysis is restricted. 
While this work represents a general proposal to compare peer-to-peer teaming with supervised teaming,
for a more accurate analysis, we need to more carefully separate the influence of the level of automation. 
To achieve this, we plan to study a more complex task domain, which also implies that the sub-goals would be more difficult to identify. 
We can design a new interface that allows the human to specify sub-goals as a supervised teaming interface, 
which can then be used to compare with a peer-to-peer teaming interface.

 
Another interesting angle is the study of human-robot interaction with one human and multiple robots.
We briefly discussed about the fan out measure. 
An interesting study is to investigate how to design teaming protocols, 
in order to increase the maximum number of robots that a human can interact with in a given scenario.

\section{Conclusions}
\label{sec:conclusion}

In this paper, we investigate the incorporation of general planning capability into robots to facilitate peer-to-peer human-robot teaming,
in which the human and robot are viewed as teammates that are physically separated.
The human and robot share the same global goal and collaborate to achieve it. 
One of the important differences between peer-to-peer teaming and supervised teaming is that,
the robot in peer-to-peer teaming can achieve the goal alone when the task information is completely specified.
However, incompleteness often exists, which implies information asymmetry.
While information asymmetry can be desirable sometimes, 
it may also lead to the robot choosing improper actions that can negatively influence the teaming performance.
We show that, in general, the human teammates prefer to work in peer-to-peer teaming.
However, our results show that peer-to-peer teaming may not reduce mental workload in short-term tasks. 
This is understandable since working with a proactive teammate may require more interactions at the beginning.
Once a mutual understanding is established, we speculate that the mental workload would be reduced.
Furthermore, we also show that the situation awareness is not significantly reduced in peer-to-peer teaming. 
This seems to suggest that providing ``recommended actions'' may also be used as a way to reduce the negative influence of automation. 
This is more broadly related to research directions in plan explanation and excuse generation.



%
\bibliographystyle{abbrv}
\bibliography{sigproc}  

\begin{thebibliography}{10}

\bibitem{briel-psp-2004}
M.~V.~D. Briel, R.~Sanchez, M.~B. Do, and S.~Kambhampati.
\newblock Effective approaches for partial satisfaction (over-subscription)
  planning.
\newblock In {\em In AAAI}, pages 562--569. AAAI Press, 2004.

\bibitem{cantrell-hri-2012}
R.~Cantrell, K.~Talamadupula, P.~W. Schermerhorn, J.~Benton, S.~Kambhampati,
  and M.~Scheutz.
\newblock Tell me when and why to do it!: run-time planner model updates via
  natural language instruction.
\newblock In {\em HRI'12}, pages 471--478, 2012.

\bibitem{casper-tc-2008}
J.~Casper and R.~Murphy.
\newblock Human-robot interactions during the robot-assisted urban search and
  rescue response at the world trade center.
\newblock {\em Systems, Man, and Cybernetics, Part B: Cybernetics, IEEE
  Transactions on}, 33(3):367--385, June 2003.

\bibitem{chen-tsmc-2011}
J.~Chen, M.~Barnes, and M.~Harper-Sciarini.
\newblock Supervisory control of multiple robots: Human-performance issues and
  user-interface design.
\newblock {\em Systems, Man, and Cybernetics, Part C: Applications and Reviews,
  IEEE Transactions on}, 41(4):435--454, July 2011.

\bibitem{christoffersen-hpce-2002}
K.~Christoffersen and D.~D. Woods.
\newblock {How to make automated systems team players}.
\newblock {\em Advances in human performance and cognitive engineering
  research}, 2:1--12, 2002.

\bibitem{Crandall-iros}
J.~Crandall and M.~Goodrich.
\newblock Characterizing efficiency of human robot interaction: a case study of
  shared-control teleoperation.
\newblock In {\em Intelligent Robots and Systems, 2002. IEEE/RSJ International
  Conference on}, volume~2, pages 1290--1295 vol.2, 2002.

\bibitem{visser-cedm-2011}
E.~de~Visser and R.~Parasuraman.
\newblock Adaptive aiding of human-robot teaming: Effects of imperfect
  automation on performance, trust, and workload.
\newblock {\em Journal of Cognitive Engineering and Decision Making},
  5(2):209--231, 2011.

\bibitem{fox-jair-2003}
M.~Fox and D.~Long.
\newblock Pddl2.1: An extension to pddl for expressing temporal planning
  domains.
\newblock {\em J. Artif. Int. Res.}, 20(1):61--124, Dec. 2003.

\bibitem{gerevini-pddl3-2006}
A.~Gerevini and D.~Long.
\newblock {Plan constraints and preferences in PDDL3}.
\newblock In {\em ICAPS Workshop on Soft Constraints and Preferences in
  Planning}, 2006.

\bibitem{goodrich-issr-2007}
M.~Goodrich, J.~Cooper, J.~Adams, C.~Humphrey, R.~Zeeman, and B.~Buss.
\newblock Using a mini-uav to support wilderness search and rescue: Practices
  for human-robot teaming.
\newblock In {\em Safety, Security and Rescue Robotics, 2007. SSRR 2007. IEEE
  International Workshop on}, pages 1--6, Sept 2007.

\bibitem{goodrich-smc-2003}
M.~Goodrich and D.~Olsen.
\newblock Seven principles of efficient human robot interaction.
\newblock In {\em Systems, Man and Cybernetics, 2003. IEEE International
  Conference on}, volume~4, pages 3942--3948 vol.4, Oct 2003.

\bibitem{Hoffmann2006507}
J.~Hoffmann and R.~I. Brafman.
\newblock Conformant planning via heuristic forward search: A new approach.
\newblock {\em Artificial Intelligence}, 170(6–7):507 -- 541, 2006.

\bibitem{kambhampati-lite-2007}
S.~Kambhampati.
\newblock Model-lite planning for the web age masses: The challenges of
  planning with incomplete and evolving domain models, 2007.

\bibitem{klein-tcm-2004}
G.~Klein, D.~D. Woods, J.~M. Bradshaw, R.~R. Hoffman, and P.~J. Feltovich.
\newblock Ten challenges for making automation a "team player" in joint
  human-agent activity.
\newblock {\em IEEE Intelligent Systems}, 19(6):91--95, Nov. 2004.

\bibitem{mcdermott-pddl-1998}
D.~Mcdermott, M.~Ghallab, A.~Howe, C.~Knoblock, A.~Ram, M.~Veloso, D.~Weld, and
  D.~Wilkins.
\newblock {PDDL - The Planning Domain Definition Language}.
\newblock Technical report, CVC TR-98-003/DCS TR-1165, Yale Center for
  Computational Vision and Control, 1998.

\bibitem{nguyen-partialp-2012}
T.~A. Nguyen, M.~Do, A.~E. Gerevini, I.~Serina, B.~Srivastava, and
  S.~Kambhampati.
\newblock Generating diverse plans to handle unknown and partially known user
  preferences.
\newblock {\em Artificial Intelligence}, 190(0):1 -- 31, 2012.

\bibitem{nickel-image-2007}
K.~Nickel and R.~Stiefelhagen.
\newblock Visual recognition of pointing gestures for human-robot interaction.
\newblock {\em Image and Vision Computing}, 25(12):1875 -- 1884, 2007.
\newblock The age of human computer interaction.

\bibitem{nourbakhsh-pc-2005}
I.~R. Nourbakhsh, K.~Sycara, M.~Koes, M.~Yong, M.~Lewis, and S.~Burion.
\newblock Human-robot teaming for search and rescue.
\newblock {\em IEEE Pervasive Computing}, 4(1):72--78, 2005.

\bibitem{parasuraman-erg-2000}
R.~Parasuraman.
\newblock Designing automation for human use: empirical studies and
  quantitative models.
\newblock {\em Ergonomics}, 43(7):931--951, 2000.
\newblock PMID: 10929828.

\bibitem{parasuraman-smc-2000}
R.~Parasuraman, T.~Sheridan, and C.~D. Wickens.
\newblock A model for types and levels of human interaction with automation.
\newblock {\em Systems, Man and Cybernetics, Part A: Systems and Humans, IEEE
  Transactions on}, 30(3):286--297, May 2000.

\bibitem{heath-hpsa-2004}
H.~A. Ruff, G.~L. Calhoun, M.~H. Draper, J.~V. Fontejon, and B.~J. Guilfoos.
\newblock Exploring automation issues in supervisory control of multiple uavs.
\newblock In {\em Proceedings of the Human Performance, Situation Awareness,
  and Automation Technology Conference}, pages 218--222, Mar. 2004.

\bibitem{ruff-ptve-2002}
H.~A. Ruff, S.~Narayanan, and M.~H. Draper.
\newblock Human interaction with levels of automation and decision-aid fidelity
  in the supervisory control of multiple simulated unmanned air vehicles.
\newblock {\em Presence: Teleoper. Virtual Environ.}, 11(4):335--351, Aug.
  2002.

\bibitem{scheutz-hri-2006}
M.~Scheutz, P.~Schermerhorn, and J.~Kramer.
\newblock The utility of affect expression in natural language interactions in
  joint human-robot tasks.
\newblock In {\em Proceedings of the 1st ACM SIGCHI/SIGART Conference on
  Human-robot Interaction}, HRI '06, pages 226--233, New York, NY, USA, 2006.
  ACM.

\bibitem{sheridan-rhfe-2005}
T.~B. Sheridan and R.~Parasuraman.
\newblock Human-automation interaction.
\newblock {\em Reviews of Human Factors and Ergonomics}, 1(1):89--129, 2005.

\bibitem{sisbot-tro-2007}
E.~Sisbot, L.~Marin-Urias, R.~Alami, and T.~Simeon.
\newblock A human aware mobile robot motion planner.
\newblock {\em Robotics, IEEE Transactions on}, 23(5):874--883, Oct 2007.

\bibitem{staudte-hri-2009}
M.~Staudte and M.~Crocker.
\newblock Visual attention in spoken human-robot interaction.
\newblock In {\em Human-Robot Interaction (HRI), 2009 4th ACM/IEEE
  International Conference on}, pages 77--84, March 2009.

\bibitem{stiefelhagen-iros-2004}
R.~Stiefelhagen, C.~Fugen, R.~Gieselmann, H.~Holzapfel, K.~Nickel, and
  A.~Waibel.
\newblock Natural human-robot interaction using speech, head pose and gestures.
\newblock In {\em IEEE/RSJ International Conference on Intelligent Robots and
  Systems}, volume~3, pages 2422--2427 vol.3, Sept 2004.

\bibitem{talamadupula-tist-2010}
K.~Talamadupula, J.~Benton, S.~Kambhampati, P.~Schermerhorn, and M.~Scheutz.
\newblock Planning for human-robot teaming in open worlds.
\newblock {\em ACM Trans. Intell. Syst. Technol.}, 1(2):14:1--14:24, Dec. 2010.

\bibitem{unhelkar-hri-2014}
V.~V. Unhelkar, H.~C. Siu, and J.~A. Shah.
\newblock Comparative performance of human and mobile robotic assistants in
  collaborative fetch-and-deliver tasks.
\newblock In {\em Proceedings of the 2014 ACM/IEEE International Conference on
  Human-robot Interaction}, HRI '14, pages 82--89, New York, NY, USA, 2014.
  ACM.

\end{thebibliography}
%
%
\end{document}